\documentclass{article} % For LaTeX2e
\usepackage{iclr2019_conference,times}
\iclrfinalcopy

\usepackage{amsmath,amsfonts,bm}

\def\figref#1{figure~\ref{#1}}
\def\Figref#1{Figure~\ref{#1}}

\def\eqref#1{equation~\ref{#1}}
\def\1{\bm{1}}

\DeclareMathAlphabet{\mathsfit}{\encodingdefault}{\sfdefault}{m}{sl}
\SetMathAlphabet{\mathsfit}{bold}{\encodingdefault}{\sfdefault}{bx}{n}

\usepackage{graphicx}
\usepackage[hidelinks=true]{hyperref}
\usepackage{multirow}

\usepackage{paralist}
\usepackage[font=small,labelfont=bf]{caption}
\usepackage{subcaption}

\newif\ifshowcomments
\showcommentstrue
\title{STCN: Stochastic Temporal Convolutional Networks}

\author{Emre Aksan \& Otmar Hilliges \\
Department of Computer Science \\
ETH Zurich, Switzerland \\
\texttt{\{emre.aksan, otmar.hilliges\}@inf.ethz.ch}
}

\definecolor{caribbeangreen}{rgb}{0.0, 0.8, 0.6}
\definecolor{cambridgeblue}{rgb}{0.64, 0.76, 0.68}
\definecolor{celadon}{rgb}{0.67, 0.88, 0.69}
\definecolor{armygreen}{rgb}{0.29, 0.33, 0.13}
\definecolor{carminered}{rgb}{1.0, 0.0, 0.22}
\definecolor{americanrose}{rgb}{1.0, 0.01, 0.24}
\definecolor{amber(sae/ece)}{rgb}{1.0, 0.49, 0.0}
\definecolor{asparagus}{rgb}{0.53, 0.66, 0.42}
\definecolor{LightCyan}{rgb}{0.88,1,1}
\definecolor{LightGreen}{rgb}{0.77,0.93,0.8}
\definecolor{LightOrange}{rgb}{0.95,.67,0.47}
\definecolor{DarkYellow}{rgb}{0.478,.478,0.0058}

\newcommand{\vecb}[1]{\textbf{#1}}
\newcommand{\vect}[1]{{#1}}
\newcommand{\expectation}{\mathbb{E}}

\newcommand{\note}[3]{{\color{#2}[#1: #3]}}
\definecolor{gold}{rgb}{0.80,.60,0}

\ifshowcomments
        \newcommand{\OH}[1]{\note{Otmar}{red}{#1}}
		\newcommand{\EA}[1]{\note{E}{blue}{#1}}
        
\else
        \newcommand{\note}[3]{\unskip}
        \newcommand{\OH}[1]{\unskip}
		\newcommand{\EA}[1]{\unskip}
        
\fi

\newcommand{\modelname}{STCN}
\newcommand{\modelnameskip}{STCN-dense}

\newcommand{\fig}{Fig.}
\newcommand{\refequ}[1] {Eq.~(\ref{#1})}

\begin{document}

\maketitle

% !TEX root = ../iclr2019_stcn.tex

\begin{abstract}
Convolutional architectures have recently been shown to be competitive on many sequence modelling tasks when compared to the de-facto standard of recurrent neural networks (RNNs), while providing computational and modeling advantages due to inherent parallelism. 
However, currently there remains a performance gap to more expressive stochastic RNN variants, especially those with several layers of dependent random variables. 
In this work, we propose stochastic temporal convolutional networks (STCNs), a novel architecture that combines the computational advantages of temporal convolutional networks (TCN) with the representational power and robustness of stochastic latent spaces. 
In particular, we propose a hierarchy of stochastic latent variables that captures temporal dependencies at different time-scales. The architecture is modular and flexible due to decoupling of deterministic and stochastic layers.
We show that the proposed architecture achieves state of the art log-likelihoods across several tasks. Finally, the model is capable of predicting high-quality synthetic samples over a long-range temporal horizon in modeling of handwritten text.
\end{abstract}
% !TEX root = ../iclr2019_stcn.tex

\section{Introduction}\label{sec:introduction}
Generative modeling of sequence data requires capturing long-term dependencies and learning of 
correlations between output variables at the same time-step. Recurrent neural networks (RNNs) and its variants have been very successful in a vast number of problem domains which rely on sequential data. Recent work in audio synthesis, language modeling and machine translation tasks \citep{dauphin2016language, van2016wavenet, dieleman2018challenge, gehring2017convolutional} has demonstrated that temporal convolutional networks (TCNs) can also achieve at least competitive performance without relying on recurrence, and hence reducing the computational cost for training. 

Both RNNs and TCNs model the joint probability distribution over sequences by decomposing the distribution over discrete time-steps. In other words, such models are trained to predict the next step, given all previous time-steps. RNNs are able to model long-term dependencies by propagating information through their deterministic hidden state, acting as an internal memory. In contrast, TCNs leverage large receptive fields by stacking many dilated convolutions, allowing them to model even longer time scales up to the entire sequence length. It is noteworthy that there is no explicit temporal dependency between the model outputs and hence the computations can be performed in parallel. The TCN architecture also introduces a temporal hierarchy: the upper layers have access to longer input sub-sequences and learn representations at a larger time scale. The local information from the lower layers is propagated through the hierarchy by means of residual and skip connections \citep{van2016wavenet, bai2018empirical}.

However, while TCN architectures have been shown to perform similar or better than standard recurrent architectures on particular tasks \citep{van2016wavenet,bai2018empirical}, there currently remains a performance gap to more recent stochastic RNN variants \citep{bayer2014learning, chung2015recurrent, fabius2014variational, fraccaro2016sequential, goyal2017z, shabanian2017variational}. Following a similar approach to stochastic RNNs, \cite{lai2018stochastic} present a significant improvement in the log-likelihood when a TCN model is coupled with latent variables, albeit at the cost of limited receptive field size.

In this work we propose a new approach for augmenting TCNs with random latent variables, that decouples deterministic and stochastic structures yet leverages the increased modeling capacity efficiently.
Motivated by the simplicity and computational advantages of TCNs and the robustness and performance of stochastic RNNs, we introduce stochastic temporal convolutional networks (\modelname) by incorporating a hierarchy of stochastic latent variables into TCNs which enables learning of representations at many timescales. However, due to the absence of an internal state in TCNs, introducing latent random variables analogously to stochastic RNNs is not feasible. Furthermore, defining conditional random variables across time-steps would result in breaking the parallelism of TCNs and is hence undesirable. 

In \modelname{} the latent random variables are arranged in correspondence to the temporal hierarchy of the TCN blocks, effectively distributing them over the various timescales (see \figref{fig:stcn-graph-colored}). Crucially, our hierarchical latent structure is designed to be a modular add-on for any temporal convolutional network architecture. Separating the deterministic and stochastic layers allows us to build \modelname{}s without requiring modifications to the base TCN architecture, and hence retains the scalability of TCNs with respect to the receptive field. This conditioning of the latent random variables via different timescales is especially effective in the case of TCNs. We show this experimentally by replacing the TCN layers with stacked LSTM cells, leading to reduced performance compared to \modelname{}. 

We propose two different inference networks. In the canonical configuration, samples from each latent variable are passed down from layer to layer and only one sample from the lowest layer is used to condition the prediction of the output. In the second configuration, called \modelnameskip, we take inspiration from recent CNN architectures \citep{huang2017densely} and utilize samples from all latent random variables via concatenation before computing the final prediction.

Our contributions can thus be summarized as:
\begin{inparaenum}[1)]
	\item We present a modular and scalable approach to augment temporal convolutional network models with effective stochastic latent variables.
	\item We empirically show that the \modelnameskip{} design prevents the model from ignoring latent variables in the upper layers \citep{zhao2017learning}. 
	\item We achieve state-of-the-art log-likelihood performance, measured by ELBO, on the IAM-OnDB, Deepwriting, TIMIT and the Blizzard datasets. 
	\item Finally we show that the quality of the synthetic samples matches the significant quantitative improvements. 
\end{inparaenum}
% !TEX root = ../iclr2019_stcn.tex

\section{Background}\label{sec:background}
\iffalse
\begin{figure}[t!]
	\centering
	%\fbox{
	\includegraphics[width=0.49\columnwidth, trim={100pt 130pt 100pt 30pt}]{fig/stcn-graph-colored.pdf}
	\hspace{20pt}
	%}
	%\fbox{
	\includegraphics[width=0.40\columnwidth, trim={5pt 280pt 300pt 50pt}]{fig/tcn-stcn-blocks.pdf}
	%}
	\caption{(\textit{Left}): The computational graph of \modelname s. The area shaded in gray denotes the inference model. The approximate posterior $q$ is conditioned on $\vecb{d}_t$ and is updated by the prior $p$ which is conditioned on the TCN representations of the previous time-step $\vecb{d}_{t-1}$. The random latent variables at the upper layers have access to a longer history while the lower layers receive more recent input steps. %The prediction is made by either using $\vect{z}_t^{1}$ or all $\vect{z}_t^{l}$'s.
	(\textit{Right}): a deterministic block $\vecb{d}_t$, consisting of $K$ Wavenet blocks \citep{van2016wavenet} (shown in inset). The dilation exponentially increases within the block.}
	\vspace{-12pt}
	\label{fig:stcn-graph-colored}
\end{figure}
\fi

\begin{figure}[t!]
 	\centering
 	\includegraphics[width=0.99\columnwidth, trim={0pt 0pt 0pt 0pt}]{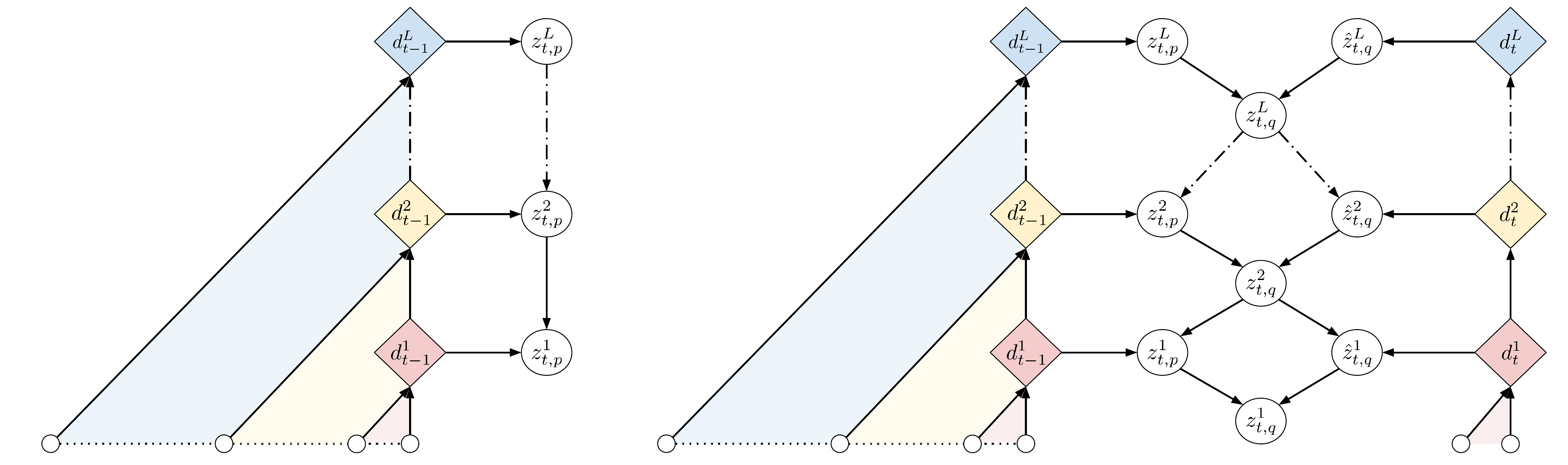}
 	\caption{The computational graph of generative (left) and  inference (right) models of \modelname. The approximate posterior $q$ is conditioned on $\vecb{d}_t$ and is updated by the prior $p$ which is conditioned on the TCN representations of the previous time-step $\vecb{d}_{t-1}$. The random latent variables at the upper layers have access to a long history while lower layers receive inputs from more recent time steps.}
 	\label{fig:stcn-graph-colored}
\end{figure}

Auto-regressive models such as RNNs and TCNs factorize the joint probability of a variable-length sequence $\vecb{x} = \{x_1, \dots, x_T\}$ as a product of conditionals as follows:
\begin{equation}
p_{\theta}(\vecb{x}) = \prod_{t=1}^{T} p_{\theta}(x_t | x_{1:t-1}) \quad ,
\label{joint-probability}
\end{equation}
where the joint distribution is parametrized by $\theta$. The prediction at each time-step is conditioned on all previous observations.
The observation model is frequently chosen to be a Gaussian or Gaussian mixture model (GMM) for real-valued data, and a categorical distribution for discrete-valued data.

\subsection{Temporal Convolutional Networks}
In TCNs the joint probabilities in \refequ{joint-probability} are parametrized by a stack of convolutional layers. \textit{Causal convolutions} are the central building block of such models and are designed to be asymmetric such that the model has no access to future information. In order to produce outputs of the same size as the input, zero-padding is applied at every layer.

In the absence of a state transition function, a large receptive field is crucial in capturing long-range dependencies. To avoid the need for vast numbers of causal convolution layers, typically \textit{dilated} convolutions are used. Exponentially increasing the dilation factor results in an exponential growth of the receptive field size with depth \citep{yu2015multi, van2016wavenet, bai2018empirical}. In this work, without loss of generality, we use the building blocks of Wavenet \citep{van2016wavenet} as gated activation units \citep{van2016conditional} have been reported to perform better.

A deterministic TCN representation $\vect{d}_{t}^l$ at time-step $t$ and layer $l$ summarizes the input sequence $\vect{x}_{1:t}$:
\begin{equation}
\vect{d}_t^{l} = \text{Conv}^{(l)}(\vect{d}_t^{l-1}, \vect{d}_{t-j}^{l-1}) \quad \text{and} \quad \vect{d}_{t}^1 = \text{Conv}^{(1)} (\vect{x_{t}}, \vect{x_{t-j}}) \quad ,
\label{tcn-deterministic-update}
\end{equation}
where the filter width is $2$ and $j$ denotes the dilation step. In our work, the stochastic variables $\vect{z^l}, l=1 \dots L$ are conditioned on TCN representations $\vect{d}^l$ that are constructed by stacking $K$ Wavenet blocks over the previous $\vect{d}^{l-1}$ (for details see \Figref{fig:network-architecture} in Appendix).

\subsection{Non-sequential Latent Variable Models}
VAEs \citep{kingma2013auto, rezende2014stochastic} introduce a latent random variable $\vecb{z}$ to learn the variations in the observed non-sequential data where the generation of the sample $\vecb{x}$ is conditioned on the latent variable $\vecb{z}$. The joint probability distribution is defined as:
\begin{equation}
	p_{\theta}({\vecb{x}, \vecb{z}}) = p_{\theta}(\vecb{x}|\vecb{z}) p_{\theta}(\vecb{z}) \quad ,
	\label{vae-non-sequential}
\end{equation}
and parametrized by $\theta$. Optimizing the marginal likelihood is intractable due to the non-linear mappings between $\vecb{z}$ and $\vecb{x}$ and the integration over $\vecb{z}$. Instead the VAE framework introduces an approximate posterior $q_{\phi}(\vecb{z}|\vecb{x})$ and optimizes a lower-bound on the marginal likelihood:
\begin{equation}
\log p_{\theta}(\vecb{x}) \geq -KL(q_{\phi}(\vecb{z}|\vecb{x}) || p_{\theta}(\vecb{z})) + \expectation_{q_{\phi}(\vecb{z}|\vecb{x})} [\log p_{\theta}(\vecb{x}|\vecb{z})] \quad ,
\label{vae-lower-bound}
\end{equation}
where $KL$ denotes the Kullback-Leibler divergence. Typically the prior $p_{\theta}(\vecb{z})$ and the approximate $q_{\phi}(\vecb{z}|\vecb{x})$ are chosen to be in simple parametric form, such as a Gaussian distribution with diagonal covariance, which allows for an analytical calculation of the $KL$-term in \refequ{vae-lower-bound}. 

\subsection{Stochastic RNNs}
An RNN captures temporal dependencies by recursively processing each input, while updating an internal state $\vect{h}_t$ at each time-step via its state-transition function:
\begin{equation}
\vect{h}_t=f^{(h)}(x_t,\vect{h}_{t-1}) \quad ,
\end{equation}
where $f^{(h)}$ is a deterministic transition function such as LSTM \citep{hochreiter1997long} or GRU \citep{cho2014learning} cells. The computation has to be sequential because $\vect{h}_t$ depends on $h_{t-1}$.

The VAE framework has been extended for sequential data, where a latent variable $z_t$ augments the RNN state $h_t$ at each sequence step. The joint distribution $p_{\theta}(\vecb{x}, \vecb{z})$ is modeled via an auto-regressive model which results in the following factorization:
\begin{equation}
p_{\theta}(\vecb{x}, \vecb{z}) = \prod_{t=1}^{T} p_{\theta} (x_t | z_{1:t}, x_{1:t-1}) p_{\theta}(z_t | x_{1:t-1}, z_{1:t-1}) \quad .
\label{vae_sequential}
\end{equation}
% VAEs define a fixed prior with $\mathcal{N}(\vecb{0}, \vecb{I})$.
In contrast to the fixed prior of VAEs, $\mathcal{N}(\vecb{0}, \vecb{I})$, sequential variants define prior distributions conditioned on the RNN hidden state $\vecb{h}$ and implicitly on the input sequence $\vecb{x}$ \citep{chung2015recurrent}.

% !TEX root = ../iclr2019_stcn.tex

\begin{figure}[t!]
	\centering
	\includegraphics[width=0.5\columnwidth, trim={0pt 0pt 0pt 0pt}]{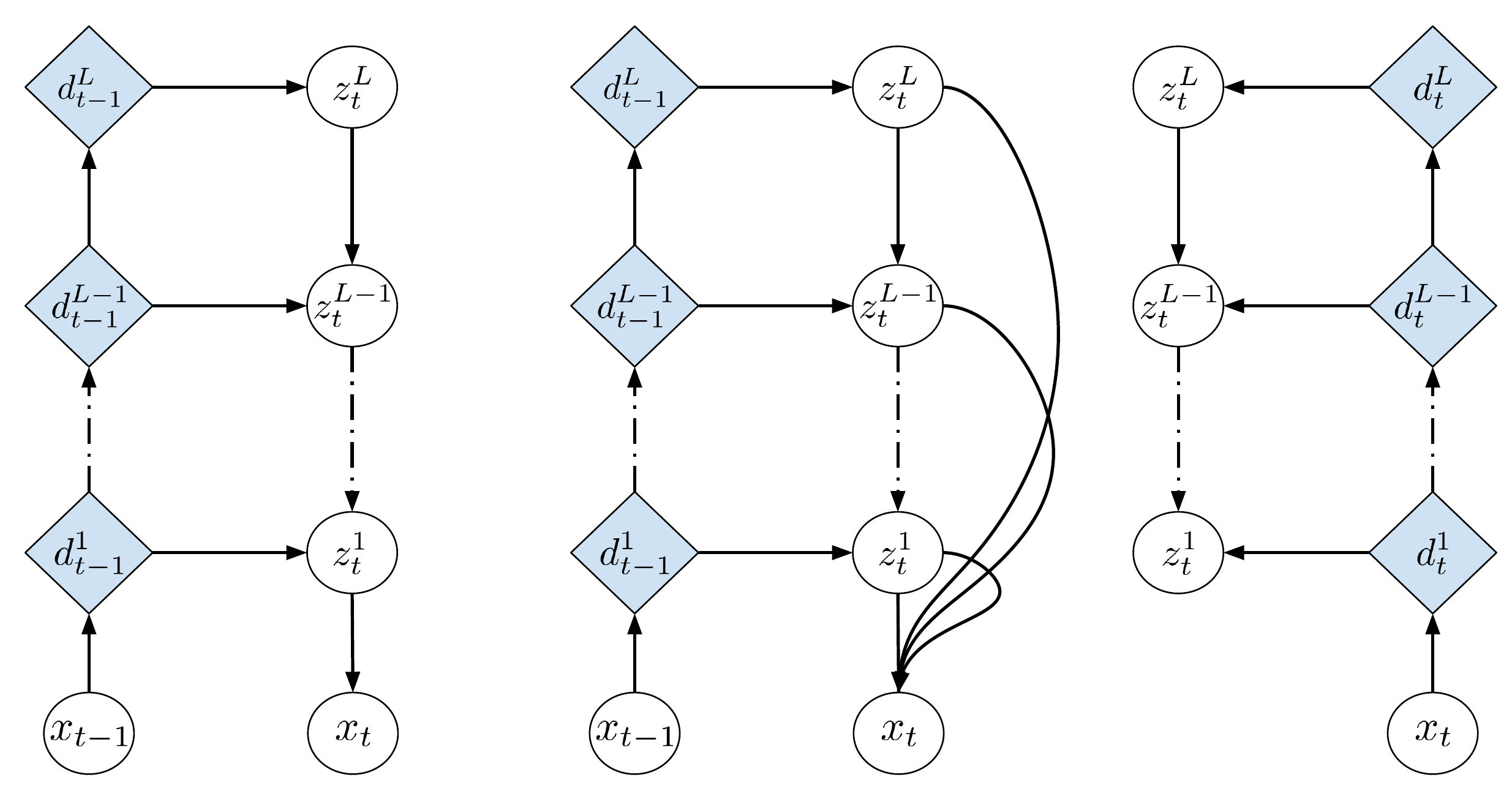}
	\caption{Graphical model view of generative models of \modelname ~(\textit{left}) and \modelnameskip ~(\textit{middle}), and the inference model (\textit{right}), which is shared by both variants. Diamonds represent the outputs of deterministic dilated convolution blocks where the dependence of $d_t$ on the past inputs is not shown for clarity (see \refequ{tcn-deterministic-update}). $\vect{x}_t$ and $\vect{z}_t$ are observable inputs and latent random variables, respectively. The generative task is to predict the next step in the sequence, given all past steps. Note that in the \modelnameskip{} variant the next step is conditioned on all latent variables $\vect{z}_t^l$ for $l = 1 \dots L$.}
	\label{fig:stcn-graphical-sketch}
\end{figure}

\section{Stochastic Temporal Convolutional Networks}\label{sec:method}
The mechanics of STCNs are related to those of VRNNs and LVAEs. Intuitively, the RNN state $\vect{h}_t$ is replaced by temporally independent TCN layers $\vect{d}_{t}^{l}$. In the absence of an internal state, we define hierarchical latent variables $\vect{z}_{t}^{l}$ that are conditioned \textit{vertically}, i.e., in the same time-step, but independent \textit{horizontally}, i.e., across time-steps. We follow a similar approach to LVAEs \citep{sonderby2016ladder} in defining the hierarchy in a \textit{top-down} fashion and in how we estimate the approximate posterior. The inference network first computes the approximate likelihood, and then this estimate is corrected by the prior, resulting in the approximate posterior. The TCN layers $\vecb{d}$ are shared between the inference and generator networks, analogous to VRNNs \citep{chung2015recurrent}.

Figure \ref{fig:stcn-graphical-sketch} depicts the proposed STCN as a graphical model. STCNs consist of two main modules: the deterministic temporal convolutional network and the stochastic latent variable hierarchy. For a given input sequence $\vecb{x} = \{\vect{x}_t\}, t=1 \dots T$ we first apply dilated convolutions over the entire sequence to compute a set of deterministic representations $\vect{d}_t^l, l= 1 \dots L$. Here, $\vect{d}_t^l$ corresponds to the output of a block of dilated convolutions at layer $l$ and time-step $t$.
The output $\vect{d}_t^l$ is then used to update a set of random latent variables $\vect{z}_t^{l}$ arranged to correspond with different time-scales.

To preserve the \textit{parallelism} of TCNs, we do not introduce an explicit dependency between different time-steps. However, we suggest that conditioning a latent variable $\vect{z}_t^{l-1}$ on the preceding variable $z_t^{l}$ implicitly introduces temporal dependencies. Importantly, the random latent variables in the upper layer have access to a larger receptive field due to its deterministic input $\vect{d}_{t-1}^l$, whereas latent random variables in lower layers are updated with different, more local information. However, the latent variable $\vect{z}_t^{l-1}$ may receive longer-range information from $\vect{z}_t^{l}$. 

The generative and inference models are jointly trained by optimizing a step-wise variational lower bound on the log-likelihood \citep{kingma2013auto, rezende2014stochastic}. In the following sections we describe these components and build up the lower-bound for a single time-step $t$.

\subsection{Generative Model}
Each sequence step $\vect{x}_t$ is generated from a set of latent variables $\vect{z}_t$, split into layers as follows:
\begin{equation}
\label{eq:p_decomposition}
	p_{\theta} (\vect{z}_t|\vect{x}_{1:t-1}) = p_{\theta} (\vect{z}_t^L | \vect{d}_{t-1}^L) \prod_{l=1}^{L-1} p_{\theta} (\vect{z}_t^l | \vect{z}_t^{l+1}, \vect{d}_{t-1}^l) \quad , \\
\end{equation}
\begin{equation}
	\text{where} \quad p_{\theta} (\vect{z}_t^l | \vect{z}_t^{l+1}, \vect{d}_{t-1}^l) = \mathcal{N} (\vect{\mu}_{t, p}^{l}, \vect{\sigma}_{t, p}^{l}) \quad \text{and} \quad
[\vect{\mu}_{t, p}^{l}, \vect{\sigma}_{t, p}^{l}] = f_p^{(l)} (\vect{z}_t^{l+1}, \vect{d}_{t-1}^l) \quad .
\end{equation}
Here the prior is modeled by a Gaussian distribution with diagonal covariance, as is common in the VAE framework. The subscript $p$ denotes items of the generative distribution. For the inference distribution we use the subscript $q$. The distributions are parameterized by a neural network $f_p^{(l)}$ and conditioned on:
\begin{inparaenum}[(1)]
	\item the $\vect{d}_{t-1}^l$ computed by the dilated convolutions from the previous time-step, and
	\item a sample from the preceding level at the same time-step $\vect{z}_t^{l+1}$.
\end{inparaenum}
Please note that at inference time we draw samples from the approximate posterior distribution $\vect{z}_t^{l+1} \sim q_{\phi} (\vect{z}_t^{l+1} | \cdot)$. The generative model, on the other hand, uses the prior $\vect{z}_t^{l+1} \sim p_{\theta}(\vect{z}_t^{l+1} | \cdot)$.

We propose two variants of the observation model. In the non-sequential scenario, the observations are defined to be conditioned on only the last latent variable in the hierarchy, i.e., $p_{\theta} (\vect{x_t} | \vect{z}_t^1)$, following \cite{sonderby2016ladder, gulrajani2016pixelvae} and \cite{rezende2014stochastic} our STCN variant uses the same observation model, allowing for an efficient optimization. However, latent units are likely to become inactive during training in this configuration \citep{burda2015importance, bowman2015generating, zhao2017learning} resulting in a loss of representational power.

The latent variables at different layers are conditioned on different contexts due to the inputs $\vect{d}_t^l$. Hence, the latent variables are expected to capture complementary aspects of the temporal context. To propagate the information all the way to the final prediction and to ensure that gradients flow through all layers, we take inspiration from \cite{huang2017densely} and directly condition the output probability on samples from \emph{all} latent variables. We call this variant of our architecture \textit{\modelnameskip}.

The final predictions are then computed by the respective observation functions:
\begin{align}
  p_{\theta} (\vect{x}_t|\vect{z}_{t}) = f^{(o)}(\vect{z}_t^1) \quad \text{and} \quad p_{\theta}^{dense} (\vect{x}_t|\vect{z}_{t}) = f^{(o)}(\vect{z}_t^{1}, \vect{z}_t^{2} \dots \vect{z}_t^{L}) \quad,
\end{align}
where $f^{(o)}$ corresponds to the output layer constructed by stacking 1D convolutions or Wavenet blocks depending on the dataset.

\subsection{Inference Model}
In the original VAE framework the inference model is defined as a bottom-up process, where the latent variables are conditioned on the stochastic layer below. Furthermore, the parameterization of the prior and approximate posterior distributions are computed separately \citep{burda2015importance, rezende2014stochastic}. In contrast, \cite{sonderby2016ladder} propose a top-down dependency structure shared across the generative and inference models. From a probabilistic point of view, the approximate Gaussian likelihood, computed bottom-up by the inference model, is combined with the Gaussian prior, computed top-down from the generative model. We follow a similar procedure in computing the approximate posterior.

First, the parameters of the approximate likelihood are computed for each stochastic layer $l$:
\begin{equation}
[\hat{\vect{\mu}}_{t, q}^{l}, \hat{\vect{\sigma}}_{t, q}^{l}] = f_q^{(l)} (\vect{z}_t^{l+1},\vect{d}_t^{l}) \quad ,
\end{equation}
followed by the downward pass, recursively computing the prior and approximate posterior by  precision-weighted addition:
\begin{equation}
\begin{aligned}
\vect{\sigma}_{t, q}^{l} &= \frac{1}{(\hat{\vect{\sigma}}_{t, q}^{l})^{-2} + (\vect{\sigma}_{t, p}^{l})^{-2}} \quad , \\
\vect{\mu}_{t, q}^{l} &= \vect{\sigma}_{t, q}^{l} (\hat{\vect{\mu}}_{t, q}^{l}(\hat{\vect{\sigma}}_{t, q}^{l})^{-2} + \vect{\mu}_{t, p}^{l}(\vect{\sigma}_{t, p}^{l})^{-2}) \quad .
\end{aligned}
\end{equation}

Finally, the approximate posterior has the same decomposition as the prior (see \refequ{eq:p_decomposition}):
\begin{equation} \label{eq:q_decomposition}
	q_{\phi} (\vect{z}_t | \vect{x}_{1:t}) = q_{\phi} (\vect{z}_t^L | \vect{d}_{t}^L) \prod_{l=1}^{L-1} q_{\phi} (\vect{z}_t^l | \vect{z}_t^{l+1}, \vect{d}_{t}^l)  \quad ,
\end{equation}
\begin{equation}
	q_{\phi} (\vect{z}_t^l | \vect{z}_t^{l+1}, \vect{d}_{t}^l) = \mathcal{N} (\vect{\mu}_{t, q}^{l}, \vect{\sigma}_{t, q}^{l}) \quad . %\text{and} \quad \vect{d}_{t}^1 = \text{Conv}^{(1)} (\vect{x_{1:t}}) \quad ,
\end{equation}
Note that the inference and generative network share the parameters of dilated convolutions $\text{Conv}^{(l)}$. 

\subsection{Learning}
The variational lower-bound on the log-likelihood at time-step $t$ can be defined as follows:
\begin{equation}
\begin{aligned}
\log p(\vect{x}_t) &\geq \expectation_{q_{\phi}(\vect{z}_t|\vect{x}_t)} [\log p_{\theta}(\vect{x}_t | \vect{z}_t)] - D_{KL}(q_{\phi}(\vect{z}_t|\vect{x}_{1:t}) || p_{\theta}(\vect{z}_t | \vect{x}_{1:t-1})) \\
&=  \expectation_{q_{\phi}(\vect{z}_t^{1} \dots \vect{z}_t^{L}|\vect{x}_t)} [\log p_{\theta}(\vect{x}_t | \vect{z}_t^{1} \dots \vect{z}_t^{L})] - D_{KL}(q_{\phi}(\vect{z}_t^{1} \dots \vect{z}_t^{L}|\vect{x}_{1:t}) || p_{\theta}(\vect{z}_t^{1} \dots \vect{z}_t^{L} | \vect{x}_{1:t-1})) \\
\mathcal{L}_t(\theta, \phi;\vect{x}_t) &= \mathcal{L}_{t}^{Recon} + \mathcal{L}_{t}^{KL} .
\end{aligned}
\end{equation}

Using the decompositions from \refequ{eq:p_decomposition} and (\ref{eq:q_decomposition}), the Kullback-Leibler divergence term becomes:
\begin{equation}
\begin{aligned}
	\mathcal{L}_{t}^{KL} = &-D_{KL}(q_{\phi}(\vect{z}_t^{L} | \vect{d}_{t}^L) || p_{\theta}(\vect{z}_t^{L} | \vect{d}_{t-1}^L)) \\ &- \sum_{l=1}^{L-1} \expectation_{q_{\phi}(\vect{z}_t^{l+1} | \cdot)} [D_{KL}(q_{\phi}(\vect{z}_t^{l} | \vect{z}_t^{l+1}, \vect{d}_{t}^l) || p_{\theta}(\vect{z}_t^{l} | \vect{z}_t^{l+1}, \vect{d}_{t-1}^l))]\quad  .
\end{aligned}
\end{equation}

The KL term is the same for the \modelname{} and \modelnameskip{} variants. The reconstruction term $\mathcal{L}_{t}^{Recon}$, however, is different. In \modelname{} we only use samples from the lowest layer of the hierarchy, whereas in \modelnameskip{} we use all latent samples in the observation model:
\begin{align}
\mathcal{L}_{t}^{Recon} &= \expectation_{q_{\phi}(\vect{z}_t^{1} \dots \vect{z}_t^{L}|\vect{x}_t)} [\log p_{\theta}(\vect{x}_t | \vect{z}_t^{1})] \quad ,\\
\mathcal{L}_{t}^{Recon-dense} &= \expectation_{q_{\phi}(\vect{z}_t^{1} \dots \vect{z}_t^{L}|\vect{x}_t)} [\log p_{\theta}(\vect{x}_t | \vect{z}_t^{1} \dots \vect{z}_t^{L}] \quad .
\end{align}

In the dense variant, samples drawn from the latent variables $z_t^l$ are carried over the dense connections. Similar to \cite{maaloe2016auxiliary}, the expectation over $z_t^l$ variables are computed by Monte Carlo sampling using the reparameterization trick \citep{kingma2013auto, rezende2014stochastic}.

Please note that the computation of $\mathcal{L}_{t}^{Recon-dense}$ does not introduce any additional computational cost. In \modelname, all latent variables have to be visited in terms of ancestral sampling in order to draw the latent sample $z_t^1$ for the observation $x_t$. Similarly in \modelnameskip, the same intermediate samples $z_t^l$ are used in the prediction of $x_t$.

One alternative option to use the latent samples could be to sum individual samples before feeding them into the observation model, i.e., $sum([z_t^1 \dots z_t^L])$, \citep{maaloe2016auxiliary}. We empirically found that this does not work well in \modelnameskip. Instead, we concatenate all samples $[z_t^1 \circ \dots \circ z_t^L]$ analogously to DenseNet \citep{huang2017densely} and \citep{kaiser2018fast}.

% !TEX root = ../iclr2019_stcn.tex

\section{Experiments}\label{sec:experiments}

\begin{figure}[b]
	\centering
	\begin{subfigure}{.25\textwidth}
		\centering
		\includegraphics[width=.99\linewidth]{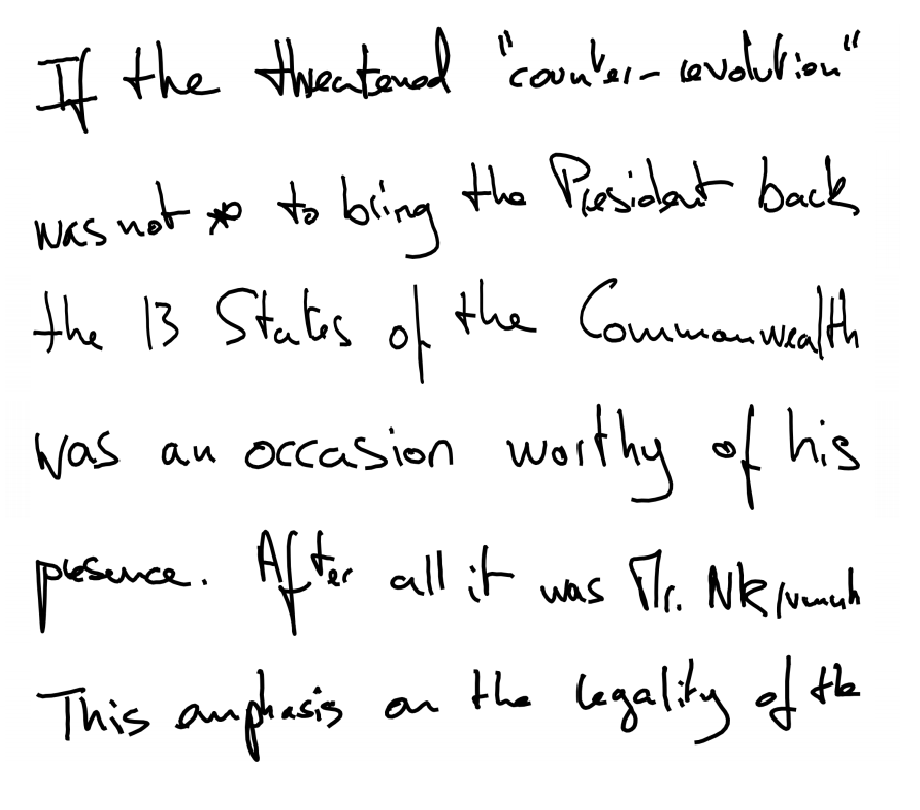}
		\caption{Ground truth}
		\label{fig:iam_gt}
	\end{subfigure}%
	\begin{subfigure}{.25\textwidth}
		\centering
		\includegraphics[width=.99\linewidth]{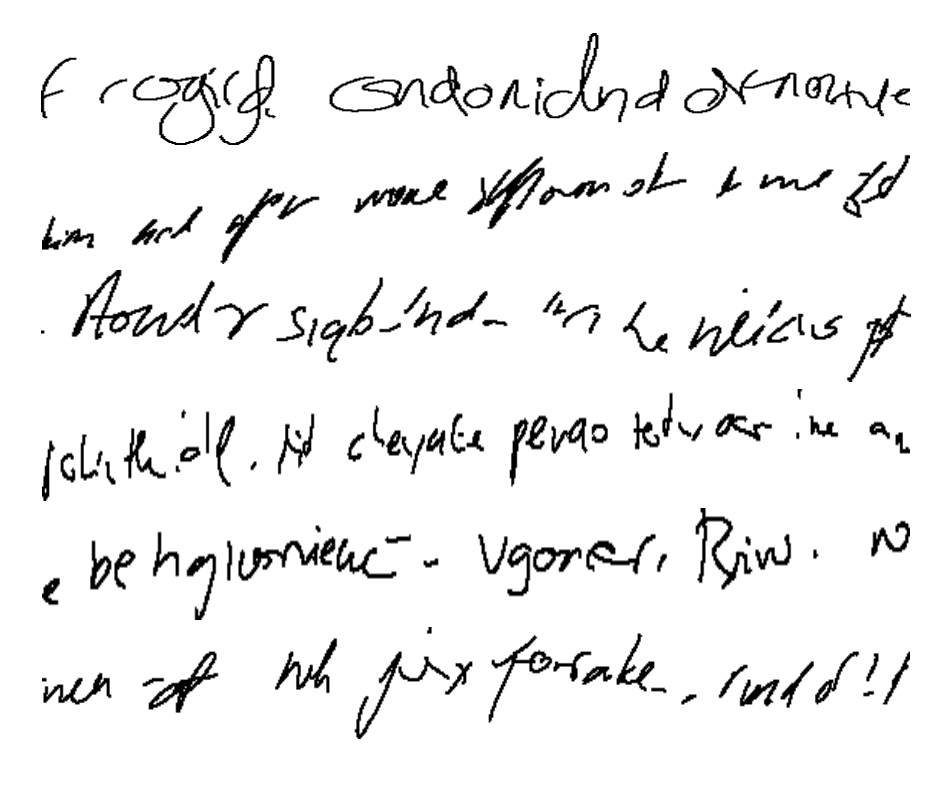}
		\caption{VRNN}
		\label{fig:iam_vrnn}
	\end{subfigure}%
	\begin{subfigure}{.24\textwidth}
		\centering
		\includegraphics[width=.99\linewidth]{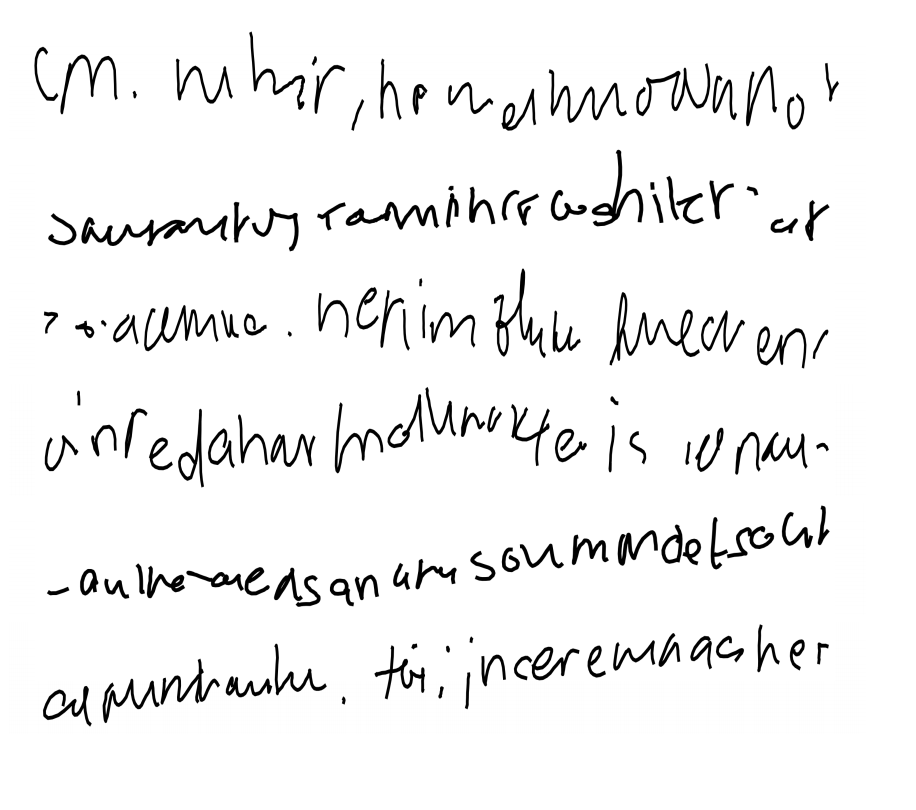}
		\caption{SwaveNet}
		\label{fig:iam_swavenet}
	\end{subfigure}%
	\begin{subfigure}{.26\textwidth}
		\centering
		\includegraphics[width=.99\linewidth]{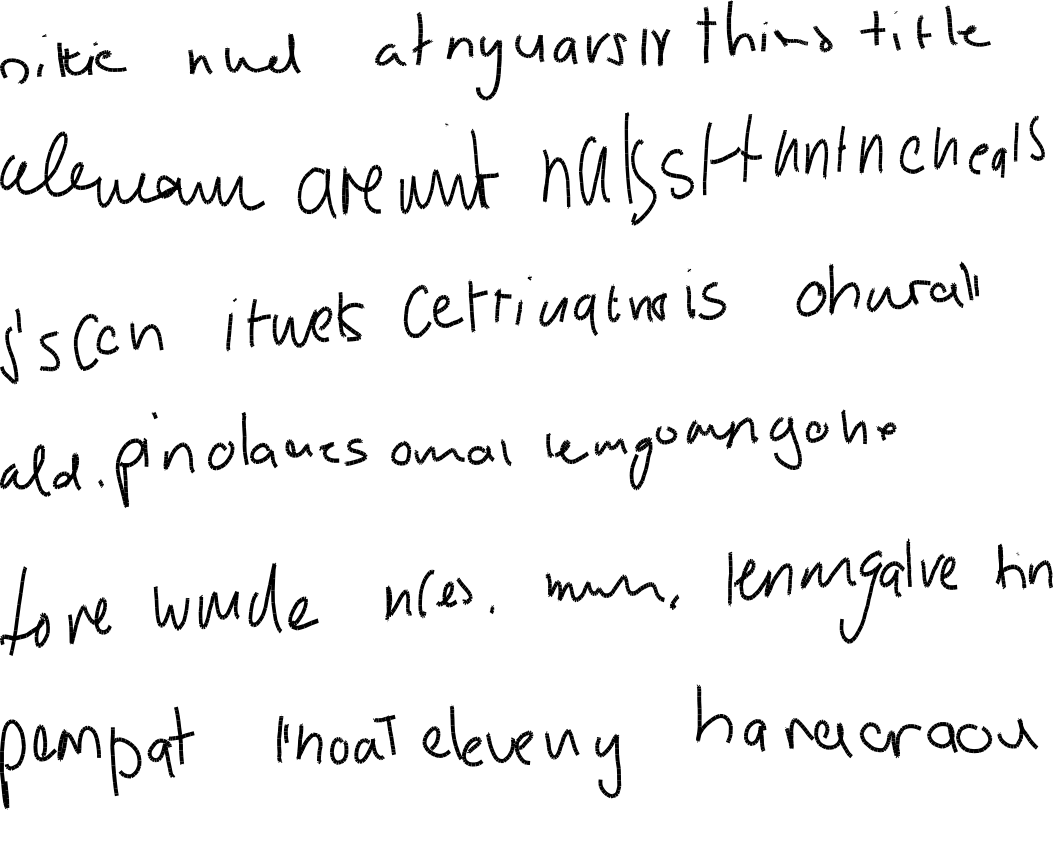}
		\caption{\modelnameskip}
		\label{fig:iam_stcn}
	\end{subfigure}%
	
	\caption{ (a) Handwriting samples from IAM-OnDB dataset. Generated samples from (b) VRNN, (c) SWaveNet and (d) our model \modelnameskip. Each line corresponds to one sample.}
	\label{fig:hw-text-samples}
\end{figure}

\begin{table}
	\centering
	\caption{Average $\log$-likelihood per sequence on TIMIT, Blizzard, IAM-OnDB and Deepwriting datasets. (Normal) and (GMM) stand for unimodal Gaussian or multi-modal Gaussian Mixture Model (GMM) as the observation model \citep{graves2013generating, chung2015recurrent}. Asterisks $^\ast$ indicate that we used our re-implementation only for the Deepwriting dataset.}
	\label{tab:all_values_combined}
	\renewcommand{\arraystretch}{1.1}
	\begin{tabular}{l c c c c}
		\hline
		Models 																		& TIMIT			& Blizzard			& IAM-OnDB			& Deepwriting \\
		\hline
		Wavenet	(GMM)																	& 30188				& 8190				& 1381				& 612 \\

		Wavenet-dense (GMM)																& 30636				& 8212				& 1380				& 642 \\

		RNN (GMM) \raisebox{2.3pt}{\tiny\cite{chung2015recurrent}}						& 26643				& 7413				& 1358				& 528 $^\ast$\\
		
		VRNN (Normal) \raisebox{2.3pt}{\tiny\cite{chung2015recurrent}} 					& $\approx$ 30235	& $\approx$ 9516	& $\approx$ 1354	& $\geq$ 495 $^\ast$\\
		
		VRNN (GMM) \raisebox{2.3pt}{\tiny\cite{chung2015recurrent}} 					& $\approx$ 29604	& $\approx$ 9392	& $\approx$ 1384	& $\geq$ 673 $^\ast$\\
		
		SRNN (Normal) \raisebox{2.3pt}{\tiny\cite{fraccaro2016sequential}}					& $\geq$ 60550		& $\geq$ 11991		& n/a		& n/a \\

		Z-forcing (Normal) \raisebox{2.3pt}{\tiny\cite{goyal2017z}}							& $\geq$ 70469		& $\geq$ 15430		& n/a		& n/a \\

		Var. Bi-LSTM (Normal) \raisebox{2.3pt}{\tiny\citet{shabanian2017variational}}	& $\geq$ 73976		& $\geq$ 17319		& n/a		& n/a \\

		SWaveNet (Normal) \raisebox{2.3pt}{\tiny\cite{lai2018stochastic}} 					& $\geq$ 72463		& $\geq$ 15708		& $\geq$ 1301		& n/a \\
		\hline
		\modelname~ (GMM)																	& $\geq$ 69195		& $\geq$ 15800			& $\geq$ 1338		& $\geq$ 605 \\
		\modelnameskip~ (GMM)																& $\geq$ 71386		& $\geq$ 16288			& $\geq$ \textbf{1796}		& $\geq$ \textbf{797} \\
		\modelnameskip-large (GMM)														& $\geq$ \textbf{77438}		& $\geq$ \textbf{17670}			& n/a				& n/a \\
		\hline
	\end{tabular}
\end{table}

We evaluate the proposed variants \modelname{} and \modelnameskip{} both quantitatively and qualitatively on modeling of digital handwritten text and speech. We compare with vanilla TCNs, RNNs, VRNNs and state-of-the art models on the corresponding tasks. 

In our experiments we use two variants of the Wavenet model: (1) the original model proposed in \citep{van2016wavenet} and (2) a variant that we augment with skip connections analogously to \modelnameskip. This additional baseline evaluates the benefit of learning \textit{multi-scale} representations in the deterministic setting. Details of the experimental setup are provided in the Appendix. Our code is available at \small \url{https://ait.ethz.ch/projects/2019/stcn/}. 

\textbf{Handwritten text:} The IAM-OnDB and Deepwriting datasets consist of digital handwriting sequences where each time-step contains real-valued $(x, y)$ pen coordinates and a binary \textit{pen-up} event. The IAM-OnDB data is split and pre-processed as done in \citep{chung2015recurrent}. \cite{aksan2018deepWriting} extend this dataset with additional samples and better pre-processing. 

Table \ref{tab:all_values_combined} reveals that again both our variants outperform the vanilla variants of TCNs and RNNs on IAM-OnDB. While the stochastic VRNN and SWaveNet are competitive wrt to the \modelname{} variant, both are outperformed by the \modelnameskip{} version. The same relative ordering is maintained on the Deepwriting dataset, indicating that the proposed architecture is robust across datasets.

\fig~\ref{fig:hw-text-samples} compares generated handwriting samples. While all models produce consistent style, our model generates more natural looking samples. Note that the spacing between words is clearly visible and most of the letters are distinguishable. 

\textbf{Speech modeling:} TIMIT and Blizzard are standard benchmark dataset in speech modeling. The models are trained and tested on $200$ dimensional real-valued amplitudes. We apply the same pre-processing as \cite{chung2015recurrent}. For this task we introduce \modelnameskip-large, with increased model capacity. Here we use 512 instead of 256 convolution filters. Note that the total number of model parameters is comparable to SWaveNet and other SOA models.

On TIMIT, \modelnameskip{} (Table \ref{tab:all_values_combined}) significantly outperforms the vanilla TCN and RNN, and stochastic models. On the Blizzard dataset, our model is marginally better than the Variational Bi-LSTM. Note that the inference models of SRNN \citep{fraccaro2016sequential}, Z-forcing \citep{goyal2017z}, and Variational Bi-LSTM \citep{shabanian2017variational} receive future information by using backward RNN cells. Similarly, SWaveNet \citep{lai2018stochastic} applies causal convolutions in the backward direction. Hence, the latent variable can be expected to model future dynamics of the sequence. In contrast, our models have only access to information up to the current time-step. These results indicate that the \modelname{} variants perform very well on the speech modeling task. 

\begin{table} [b]
	\centering
	\caption{KL-loss per latent variable computed over the entire test split. KL5 corresponds to the KL-loss of the top-most latent variable.}
	\renewcommand{\arraystretch}{1.2}
	\begin{tabular}{l||c c| c c c c c}
		\hline		
		Dataset (Model) 								& ELBO		& KL		& KL1		&KL2		&KL3		&KL4		&KL5\\
		\hline
		IAM-OnDB ({\tiny\modelnameskip})						& $\geq$ 1796.3		&1653.9		&17.9		&1287.4		&305.3		&41.0		&2.4\\
		IAM-OnDB ({\tiny\modelname})							& $\geq$ 1339.2		&964.2		&846.0		&105.2		&12.9		&0.1		&0.0\\
		TIMIT ({\tiny\modelnameskip})							& $\geq$ 71385.9	&22297.5	&16113.0	&5641.6		&529.0		&8.3		&5.7\\
		TIMIT ({\tiny\modelname})								& $\geq$ 69194.9	&23118.3	&22275.5	&487.2		&355.5		&0.0		&0.0\\
		\hline
	\end{tabular}
	\label{tab:all_kld}
	%\vspace{-10pt}
\end{table}

\textbf{Latent Space Analysis:} 
\cite{zhao2017learning} observe that in hierarchical latent variable models the upper layers have a tendency to become inactive, indicated by a low KL loss \citep{sonderby2016ladder, dieng2018avoiding}. Table \ref{tab:all_kld} shows the KL loss per latent variable and the corresponding log-likelihood measured by ELBO in our models. 
Across the datasets it can be observed that our models make use of many of the latent variables which may explain the strong performance across tasks in terms of log-likelihoods. Note that \modelname{} uses a standard hierarchical structure. However, individual latent variables have different information context due to the corresponding TCN block's receptive field. This observation suggests that the proposed combination of TCNs and stochastic variables is indeed effective. Furthermore, in \modelname{} we see a similar utilization  pattern of the $z$ variables across tasks, whereas \modelnameskip{} may have more flexibility in modeling the temporal dependencies within the data due to its dense connections to the output layer.

\textbf{Replacing TCN with RNN:} 
To better understand potential symergies between dilated CNNs and the proposed latent variable hierarchy, we perform an ablation study, isolating the effect of TCNs and the latent space. To this end the deterministic TCN blocks are replaced with LSTM cells by keeping the latent structure intact. We dub this condition LadderRNN. We use the TIMIT and IAM-OnDB datasets for evaluation. Table \ref{tab:ladderrnn} summarizes performance measured by the ELBO.

The most direct translation of the the STCN architecture into an RNN counterpart has 25 stacked LSTM cells with 256 units each. Similar to STCN, we use 5 stochastic layers (see Appendix \ref{sec:experiment_setup}). Note that stacking this many LSTM cells is unusual and resulted in instabilities during training. Hence, the performance is similar to vanilla RNNs. The second LadderRNN configuration uses 5 stacked LSTM cells with 512 units and a one-to-one mapping with the stochastic layers. On the TIMIT dataset, all LadderRNN configurations show a significant improvement. We also observe a pattern of improvement with densely connected latent variables.

This experiments shows that the proposed modular latent variable design does allow for the usage of different building blocks. Even when attached to LSTM cells, it boosts the log-likelihood performance (see 5x512-LadderRNN), in particular when used with dense connections. However, the empirical results suggest that the densely connected latent hierarchy interacts particularly well with dilated CNNs. We suggest this is due to the hierarchical nature on both sides of the architecture. On both datasets \modelname~ models achieved the best performance and significantly improve with dense connections. This supports our contribution of a latent variable hierarchy, which models different aspects of information from the input time-series. 

\begin{table}[h]
	\centering
	\caption{ELBO of LadderRNN and STCN models using the same latent space configuration. The prefix of a model entries denote the number of RNN or TCN layers and unit size per layer. Models have similar number of trainable parameters.}
	\renewcommand{\arraystretch}{1.1}
	\begin{tabular}{l c c}
		\hline
		Models & TIMIT & IAM-OnDB \\ 
		\hline
		25x256-LadderRNN (Normal)       & $\geq$ 28207 & $\geq$ 1305     \\ 
		25x256-LadderRNN-dense (Normal) & $\geq$ 27413 & $\geq$ 1278     \\ 
		25x256-LadderRNN (GMM)          & $\geq$ 24839 & $\geq$ 1381     \\ 
		25x256-LadderRNN-dense (GMM)    & $\geq$ 26240 & $\geq$ 1377     \\ \hline
		5x512-LadderRNN (Normal)        & $\geq$ 49770 & $\geq$ 1299     \\ 
		5x512-LadderRNN-dense (Normal)  & $\geq$ 48612 & $\geq$ 1374     \\ 
		5x512-LadderRNN (GMM)           & $\geq$ 47179 & $\geq$ 1359     \\ 
		5x512-LadderRNN-dense (GMM)     & $\geq$ 50113 & $\geq$ 1581     \\ \hline
		25x256-STCN (Normal)            & $\geq$ 64913 & $\geq$ 1327     \\ 
		25x256-STCN-dense (Normal)      & $\geq$ 70294 & $\geq$ 1729     \\ 
		25x256-STCN (GMM)               & $\geq$ 69195 & $\geq$ 1339     \\ 
		25x256-STCN-dense (GMM)         & $\geq$ \textbf{71386} & $\geq$ \textbf{1796}     \\
		\hline
	\end{tabular}
	\label{tab:ladderrnn}
\end{table}

% !TEX root = ../iclr2019_stcn.tex

\section{Related Work}\label{sec:related}
% Non-sequential latent variable models.
\cite{rezende2014stochastic} propose Deep Latent Gaussian Models (DLGM) and \cite{sonderby2016ladder} propose the Ladder Variational Autoencoder (LVAE). In both models the latent variables are hierarchically defined and conditioned on the preceding stochastic layer. LVAEs improve upon DLGMs via implementation of a top-down hierarchy both in the generative and inference model. The approximate posterior is computed via a precision-weighted update of the approximate likelihood (i.e., the inference model) and prior (i.e., the generative model).  Similarly, the PixelVAE \citep{gulrajani2016pixelvae} incorporates a hierarchical latent space decomposition and uses an autoregressive decoder. \cite{zhao2017learning} show under mild conditions that straightforward stacking of latent variables (as is done e.g. in LVAE and PixelVAE) can be ineffective, because the latent variables that are not directly conditioned on the observation variable become inactive.

Due to the nature of the sequential problem domain, our approach differs in the crucial aspects that \modelname{}s use dynamic, i.e., conditional, priors \citep{chung2015recurrent} at every level. Moreover, the hierarchy is not only implicitly defined by the network architecture but also explicitly defined by the information content, i.e., receptive field size. \cite{dieng2018avoiding} both theoretically and empirically show that using skip connections from the latent variable to every layer of the decoder increases mutual information between the latent and observation variables. Similar to \cite{dieng2018avoiding} in \modelnameskip, we introduce skip connections from all latent variables to the output.  In \modelname{} the model is expected to encode and propagate the information through its hierarchy.

%Autoregressive distributions in the decoder of VAE architectures. 
\cite{yang2017improved} suggest using autoregressive TCN decoders to remedy the posterior collapse problem observed in language modeling with LSTM decoders \citep{bowman2015generating}. \cite{van2017neural} and \cite{dieleman2018challenge} use TCN decoders conditioned on discrete latent variables to model audio signals.

% Sequential latent variable models.
Stochastic RNN architectures mostly vary in the way they employ the latent variable and parametrize the approximate posterior for variational inference. \cite{chung2015recurrent} and \cite{bayer2014learning} use the latent random variable to capture high-level information causing the variability observed in sequential data. Particularly \cite{chung2015recurrent} shows that using a conditional prior rather than a standard Gaussian distribution is very effective in sequence modeling. In \citep{fraccaro2016sequential, goyal2017z, shabanian2017variational}, the inference model, i.e., the approximate posterior, receives both the past and future summaries of the sequence from the hidden states of forward and backward RNN cells. The KL-divergence term in the objective enforces the model to learn predictive latent variables in order to capture the future states of the sequence.

\cite{lai2018stochastic}'s SWaveNet is most closely related to ours. SWaveNet also introduces latent variables into TCNs. However, in SWaveNet the deterministic and stochastic units are coupled which may prevent stacking of larger numbers of TCN blocks. Since the number of stacked dilated convolutions determines the receptive field size, this directly correlates with the model capacity. For example, the performance of SWaveNet on the IAM-OnDB dataset degrades after stacking more than $3$ stochastic layers \citep{lai2018stochastic}, limiting the model to a small receptive field. In contrast, we aim to preserve the flexibility of stacking dilated convolutions in the base TCN. In \modelname{}s, the deterministic TCN units do not have any dependency on the stochastic variables (see \Figref{fig:stcn-graph-colored}) and the ratio of stochastic to deterministic units can be adjusted, depending on the task.
% !TEX root = ../iclr2019_stcn.tex

\section{Conclusion}\label{sec:conclusion}
In this paper we proposed \modelname s, a novel auto-regressive model, combining the computational benefits of convolutional architectures and expressiveness of hierarchical stochastic latent spaces. We have shown the effectivness of the approach across several sequence modelling tasks and datasets. The proposed models are trained via optimization of the ELBO objective. Tighter lower bounds such as IWAE \citep{burda2015importance} or FIVO \citep{maddison2017filtering} may further improve modeling performance. We leave this for future work.

\section*{Acknowledgements}
This work was supported in parts by the ERC grant OPTINT (StG-2016-717054). We gratefully acknowledge the support of NVIDIA Corporation with the donation of the Titan Xp GPU used for this research.

\clearpage
\bibliography{iclr2019_stcn}
\bibliographystyle{iclr2019_conference}
\clearpage
% !TEX root = ../iclr2019_stcn.tex

\section{Appendix}
\subsection{Network Details}
\label{sec:experiment_setup}

\begin{figure} [h]
	\centering
	\includegraphics[width=0.99\textwidth, trim={0pt 0pt 0pt 0pt}]{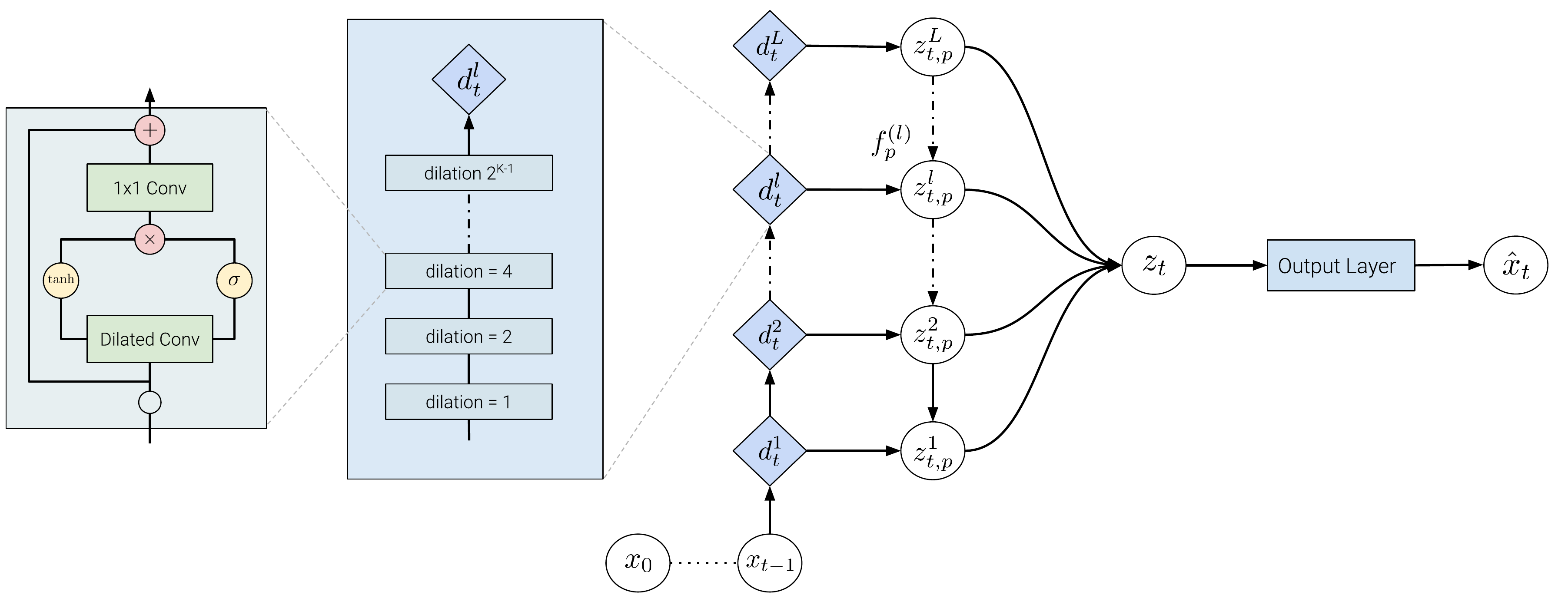}
	\caption{Generative model of \modelnameskip{} architecture. Building blocks are highlighted. Note that the dependence of $d_t^l, l=1 \cdots L$ on past inputs is not visualized for clarity.}
	\label{fig:network-architecture}
\end{figure}

The network architecture of the proposed model is illustrated in \fig~ \ref{fig:network-architecture}. We make only a small modification to the vanilla Wavenet architecture. Instead of using skip connections from Wavenet blocks, we only use the latent sample $z_t$ in order to make a prediction of $x_t$. In \modelnameskip{} configuration, $z_t$ is the concatenation of all latent variables in the hierarchy, i.e., $z_t = [z_t^1 \circ \dots \circ z_t^L]$, whereas in \modelname{} only $z_t^1$ is fed to the output layer.

Each stochastic latent variable $z_t^l$ (except the top-most $z_t^L$) is conditioned on a deterministic TCN representation $d_t^l$ and the preceding random variable $z_t^{l+1}$. The latent variables are calculated by using the latent layers $f_{p}^{(l)}$ or $f_{q}^{(l)}$ which are neural networks.

We do not define a latent variable per TCN layer. Instead, the stochastic layers are uniformly distributed where each random variable is conditioned on a number of stacked TCN layers $d_t^l$. We stack $K$ Wavenet blocks (see \figref{fig:network-architecture} left) with exponentially increasing dilation size. 

\textbf{Observation Model:} We use Normal or GMM distributions with 20 components to model real-valued data. All Gaussian distributions have diagonal covariance matrix.

\textbf{Output layer $f^{(o)}$:} For the IAM-OnDB and Deepwriting datasets we use 1D convolutions with ReLU nonlinearity. We stack 5 of these layers with 256 filters and filter size 1.

For TIMIT and Blizzard datasets Wavenet blocks in the output layer perform significantly better. We stack 5 Wavenet blocks with dilation size 1. For each convolution operation in the block we use 256 filters. The filter size of the dilated convolution is set to 2. The \modelnameskip-large model is constructed by using 512 filters instead of 256.

\textbf{TCN blocks $d_t^l$:} The number of Wavenet blocks is usually determined by the desired receptive field size. 
\begin{itemize}
\item For the handwriting datasets $K=6$ and $L=5$. In total we have 30 Wavenet blocks where each convolution operation has 256 filters with size 2.
\item For speech datasets $K=5$ and $L=5$. In total we have 25 Wavenet blocks where each convolution operation has 256 filters with size 2. The large model configuration uses 512 filters.
\end{itemize}

\textbf{Latent layers $f_{p}^{(l)}$ and $f_{q}^{(l)}$}: The number of stochastic layers per task is given by $L$. We used $[32, 16, 8, 5, 2]$ dimensional latent variables for the handwriting tasks. It is $[256, 128, 64, 32, 16]$ for speech datasets. Note that the first entry of the list corresponds to $z^1$.

The mean and sigma parameters of the Normal distributions modeling the latent variables are calculated by the $f_{p}^{(l)}$ and $f_{q}^{(l)}$ networks. We stack 2 1D convolutions with ReLU nonlinearity and filter size 1. The number of filters are the same as the number of Wavenet block filters for the corresponding task. 

Finally, we clamped the latent sigma predictions between $0.001$ and $5$.

\subsection{Training Details}
In all \modelname{} experiments we applied KL annealing. In all tasks, the weight of the KL term is initialized with 0 and increased by $1\times e^{-4}$ at every step until it reaches $1$.

The batch size was $20$ for all datasets except for Blizzard where it was $128$.

We use the ADAM optimizer with its default parameters and exponentially decay the learning rate. For the handwriting datasets the learning rate was initialized with $5\times e^{-4}$ and followed a decay rate of $0.94$ over $1000$ decay steps. On the speech datasets it was initialized with $1\times e^{-3}$ and decayed with a rate of $0.98$. We applied early stopping by measuring the ELBO performance on the validation splits.

We implement \modelname{} models in Tensorflow \citep{tensorflow}. Our code and models achieving the SOA results are available at \small \url{https://ait.ethz.ch/projects/2019/stcn/}.

\subsection{Detailed Results}
Here we provide the extended results table with Normal observation model entries for available models.
\begin{table}[h]
	\centering
	\caption{Average $\log$-likelihood per sequence on TIMIT, Blizzard, IAM-OnDB and Deepwriting datasets. (Normal) and (GMM) stand for unimodal Gaussian or multi-modal Gaussian Mixture Model (GMM) as the observation model \citep{graves2013generating, chung2015recurrent}. Asterisks $^\ast$ indicate that we used our re-implementation only for the Deepwriting dataset.}
	
	\renewcommand{\arraystretch}{1.1}
	\begin{tabular}{l c c c c}
		\hline		
		Models 																				& TIMIT				& Blizzard			& IAM-OnDB			& Deepwriting \\
		\hline

		Wavenet	(Normal)																	& -7443				& 3784				& 1053				& 337 \\

		Wavenet	(GMM)																		& 30188				& 8190				& 1381				& 612 \\

		Wavenet-dense (Normal)																& -8579				& 3712				& 1030				& 323 \\

		Wavenet-dense (GMM)																	& 30636				& 8212				& 1380				& 642 \\

		RNN (Normal) \raisebox{2.3pt}{\tiny\cite{chung2015recurrent}}						& -1900				& 3539				& 1016				& 363 $^\ast$\\

		RNN (GMM) \raisebox{2.3pt}{\tiny\cite{chung2015recurrent}}							& 26643				& 7413				& 1358				& 528 $^\ast$\\

		VRNN (Normal)\raisebox{2.3pt}{\tiny\cite{chung2015recurrent}} 						& $\approx$ 30235	& $\approx$ 9516	& $\approx$ 1354	& $\geq$ 495 $^\ast$\\

		VRNN (GMM) \raisebox{2.3pt}{\tiny\cite{chung2015recurrent}} 						& $\approx$ 29604	& $\approx$ 9392	& $\approx$ 1384	& $\geq$ 673 $^\ast$\\

		SRNN (Normal) \raisebox{2.3pt}{\tiny\cite{fraccaro2016sequential}}					& $\geq$ 60550		& $\geq$ 11991		& n/a		& n/a \\

		Z-forcing (Normal)\raisebox{2.3pt}{\tiny\cite{goyal2017z}}							& $\geq$ 70469		& $\geq$ 15430		& n/a		& n/a \\

		Var. Bi-LSTM (Normal)\raisebox{2.3pt}{\tiny\cite{shabanian2017variational}}	& $\geq$ 73976		& $\geq$ 17319		& n/a		& n/a \\

		SWaveNet (Normal)\raisebox{2.3pt}{\tiny\cite{lai2018stochastic}} 					& $\geq$ 72463		& $\geq$ 15708		& $\geq$ 1301		& n/a \\
		\hline
		\modelname (Normal)																	& $\geq$ 64913		& $\geq$ 13273			& $\geq$ 1327		& $\geq$ 575 \\

		\modelname (GMM)																	& $\geq$ 69195		& $\geq$ 15800			& $\geq$ 1338		& $\geq$ 605 \\

		\modelnameskip (Normal)																& $\geq$ 70294		& $\geq$ 15950			& $\geq$ 1729		& $\geq$ 740 \\

		\modelnameskip (GMM)																& $\geq$ 71386		& $\geq$ 16288			& $\geq$ \textbf{1796}		& $\geq$ \textbf{797} \\

		\modelnameskip-large (GMM)															& $\geq$ \textbf{77438}		& $\geq$ \textbf{17670}			& n/a				& n/a \\
		\hline
	\end{tabular}
	\label{tab:nll_big_table}
\end{table}

\end{document}